\documentclass[11pt]{article}

\usepackage[preprint]{acl}

\usepackage{times}
\usepackage{latexsym}
\usepackage[T1]{fontenc}
\usepackage[utf8]{inputenc}
\usepackage{microtype}
\usepackage{inconsolata}
\usepackage{graphicx}
\usepackage{booktabs}
\usepackage{multirow}
\usepackage{amsmath}
\usepackage{amssymb}
\usepackage[table]{xcolor}
\usepackage{colortbl}
\usepackage{tabularx}

\usepackage{tcolorbox}
\usepackage{listings}
\tcbuselibrary{breakable, skins, listings}

\lstdefinestyle{promptlst}{
  basicstyle=\ttfamily\footnotesize,
  breaklines=true,
  breakatwhitespace=false,
  columns=flexible,
  keepspaces=true,
  showstringspaces=false,
}

\newtcblisting{promptbox}[1]{%
  enhanced,
  breakable,
  colback=gray!8,
  colframe=black!75,
  colbacktitle=black!75,
  coltitle=white,
  fonttitle=\bfseries\scshape\small,
  title={#1},
  arc=1pt,
  boxrule=0.4pt,
  left=6pt, right=6pt, top=4pt, bottom=4pt,
  listing only,
  listing options={style=promptlst},
}

\title{ReM-MoA: Reasoning Memory Sustains Mixture-of-Agents Scaling}

\author{
  Heng Ping\textsuperscript{1,*}, Arijit Bhattacharjee\textsuperscript{2}, Peiyu Zhang\textsuperscript{1}, Shixuan Li\textsuperscript{1}, Wei Yang\textsuperscript{1}, \\
  \textbf{Ali Jannesari\textsuperscript{2}}, \textbf{Nesreen Ahmed\textsuperscript{3}}, \textbf{Paul Bogdan\textsuperscript{1,*}} \\
  \textsuperscript{1}University of Southern California;
  \textsuperscript{2}Iowa State University;
  \textsuperscript{3}Cisco AI Research \\
  \textsuperscript{*}\textbf{Correspondence:} hping@usc.edu, pbogdan@usc.edu
}

\begin{document}
\maketitle
\begin{abstract}

Mixture-of-Agents (MoA) architectures improve inference-time scaling by organizing multiple LLM agents into layered reasoning pipelines. However, existing MoA variants fail to sustain gains as depth increases, exhibiting degradation, early plateauing, or saturation. 
We propose ReM-MoA, a memory-augmented MoA framework that sustains scaling through two mechanisms: (1) a Ranked Reasoning Memory that persistently stores and ranks reasoning traces from all layers using a comparative Reviewer Agent, and (2) a Curated Diversified Memory Routing scheme that exposes different agents to distinct combinations of successful and failed traces, preserving exploration diversity while propagating high-quality reasoning. We further introduce an optional multi-domain Reviewer distillation pipeline that improves ranking quality through frontier-model supervision. Across five reasoning benchmarks spanning math, formal logic, code, knowledge, and commonsense, ReM-MoA consistently outperforms prior MoA variants across both depth and width scaling, and its advantage widens with depth, establishing 
structured cross-layer reasoning memory as a key missing 
mechanism for scalable multi-agent inference.
\end{abstract}

\section{Introduction}
\label{sec:intro}
Inference-time computation has emerged as a complementary scaling 
axis to parameter scaling for large language 
models~\citep{snell2024scaling}. Among inference-time strategies, 
multi-LLM collaboration leverages collective reasoning by aggregating 
outputs from multiple agents, with paradigms ranging from majority 
voting~\citep{wang2022selfconsistency} to multi-agent 
debate~\citep{du2024improving, liang2024encouraging}. A more recent 
paradigm, \textbf{Mixture-of-Agents 
(MoA)}~\citep{wang2025mixture}, organizes $N$ proposer agents into 
$L$ stacked layers followed by a final aggregator, enabling 
iterative refinement across depth. In principle, increasing depth 
and width should provide a natural path toward scalable collective 
reasoning. In practice, however, existing MoA systems fail to 
sustain gains as architectures deepen, often degrading, plateauing, 
or saturating after only a few layers.

Existing MoA variants attempt to sustain scaling through three broad strategies. The first strategy, 
\emph{information consolidation}, curates the inputs of each layer 
to reduce redundancy, exemplified by RMoA~\citep{xie2025rmoa}, which 
selects a diverse subset of prior outputs and forwards only the 
inter-layer residuals between adjacent layers. The second, 
\emph{inter-agent interaction}, exemplified by 
AttentionMoA~\citep{wen2026attention}, enables agents to refine each 
other within a layer through natural-language attention. The third, 
\emph{quality-guided selection}, exemplified by 
VeriMoA~\citep{ping2025verimoa}, maintains a global cache of the 
highest-scoring solutions but requires domain-specific verifiers such 
as hardware description language simulators, limiting its 
applicability to specialized tasks.


While these methods differ architecturally, none sustains performance as MoA depth increases. 
Three mechanisms drive this scaling collapse~\citep{xie2025rmoa,wen2026attention}. First, raw inter-layer concatenation propagates both useful reasoning and accumulated errors, allowing mistakes to compound 
across depth. Second, architectures with only local inter-layer communication fail to preserve strong reasoning trajectories generated earlier in the hierarchy. Third, as agents consume 
increasingly similar accumulated context, exploration diversity collapses and agents converge toward redundant solution paths, undermining the core benefit of multi-agent reasoning~\citep{zhang2025diversity, li2025rethinking}. More importantly, retaining memory alone is insufficient: exposing all agents to identical accumulated reasoning can itself accelerate convergence toward repetitive trajectories. Sustained MoA scaling therefore requires simultaneously preserving high-quality reasoning across layers while maintaining diversity among agents within each layer.

To address these scaling failures, we propose \textbf{ReM-MoA}, a 
Mixture-of-Agents framework centered on a \emph{Ranked Reasoning 
Memory} that stores reasoning traces produced across 
the layers. Unlike episodic or 
retrieval-based memory systems that accumulate experience across 
tasks, ReM-MoA maintains a structured cross-layer reasoning memory 
within a single problem instance, enabling deeper layers to revisit prior reasoning paths. At each layer, proposer agents generate reasoning 
traces, and a \emph{Reviewer Agent}~\citep{zheng2023judging} comparatively evaluates these traces 
by assigning scores together with targeted rationales. The ranked 
traces and rationale are then stored in memory~\citep{ouyang2025reasoningbank}. Subsequent layers consult this memory through a 
\emph{Curated Diversified Memory Routing} scheme that exposes 
different agents to distinct combinations of high-quality and 
low-quality reasoning trajectories, providing contrastive reasoning 
signals while preserving exploration diversity.

ReM-MoA therefore treats MoA scaling as 
a joint quality-diversity preservation problem rather than solely an 
aggregation or interaction problem. As an optional enhancement, the 
Reviewer Agent can be further improved through frontier-model 
distillation~\citep{hinton2015distilling}, where a strong teacher 
model supervises a smaller reviewer fine-tuned using 
parameter-efficient LoRA~\citep{hu2022lora}. Multi-domain training 
produces a general-purpose Reviewer that transfers across reasoning 
tasks. 
Our contributions are as follows:

\begin{itemize}

    \item We empirically characterize three scaling failure 
    modes in existing MoA systems: degradation under 
    raw concatenation, early plateauing under local residual 
    propagation, and diversity collapse under intra-layer 
    interaction mechanisms. These observations motivate structured 
    cross-layer reasoning memory as a critical mechanism for 
    sustained MoA scaling.

    \item We propose \textbf{ReM-MoA}, which sustains MoA scaling 
    through a Ranked Reasoning Memory and a Curated Diversified 
    Memory Routing scheme. At each layer, a comparative Reviewer Agent ranks reasoning traces with rationales, and subsequent agents 
    receive distinct combinations of successful and failed 
    traces, jointly preserving quality and diversity. An optional frontier-model distillation pipeline further upgrades the Reviewer Agent through multi-domain training.

    \item Across five reasoning benchmarks (math, formal logic, code, 
  knowledge, and commonsense), ReM-MoA consistently outperforms prior MoA variants 
    across all tested depth and width configurations, with the 
    performance gap widening rather than diminishing as the 
    architecture scales.

\end{itemize}

\section{Related Work}
\label{sec:related}

\paragraph{Mixture-of-Agents variants.}
Building on the original MoA framework~\citep{wang2025mixture}, 
several variants extend the three paradigms from 
Section~\ref{sec:intro}. Within \emph{information 
consolidation}, beyond RMoA~\citep{xie2025rmoa}, 
SMoA~\citep{li2025smoa} sparsifies agent interactions via 
dedicated judge and moderator agents, while 
Self-MoA~\citep{li2025rethinking} instead argues quality 
dominates diversity, aggregating samples from a single 
strong model. Within \emph{inter-agent interaction}, beyond 
AttentionMoA~\citep{wen2026attention}, 
DEI~\citep{zhang2025diversity} ensembles specialist agents 
with complementary expertise. The \emph{quality-guided} 
paradigm is exemplified by VeriMoA~\citep{ping2025verimoa} 
and related agentic frameworks for hardware description 
language generation~\citep{ping2025hdlcore, ping2026coevo, ping2026poet}, which rely on domain-specific feedback such as hardware simulators. No prior MoA variant maintains a structured cross-layer memory of reasoning traces accessible to subsequent layers.

\paragraph{Memory in LLM agents.}
Explicit memory has been used for LLM agent 
self-improvement: Reflexion~\citep{shinn2023reflexion} stores 
verbal self-reflection from past failures, 
ReasoningBank~\citep{ouyang2025reasoningbank} distills 
reasoning strategies from successful and failed 
trajectories across tasks, and 
AgentAuditor~\citep{luo2026agentauditor} augments LLM 
evaluators with retrieval-augmented examples for safety 
judgment. Our Ranked Reasoning Memory instead operates 
within a single problem across MoA depth, holding ranked 
traces from each layer for subsequent layers to consult.

\paragraph{LLM-as-Judge and judge distillation.}
LLM-as-Judge~\citep{zheng2023judging} has emerged as a 
scalable proxy for human preference, with pairwise 
comparison generally more reliable than independent 
scoring~\citep{yang2026auditing}. JudgeLM~\citep{zhu2025judgelm} distills 
frontier-model judgments into open-source judges via 
knowledge distillation~\citep{hinton2015distilling} and 
parameter-efficient fine-tuning~\citep{hu2022lora}. Our 
Reviewer Agent extends this with a layer-wise comparative 
ranking that feeds a downstream reasoning memory.

\begin{figure*}[t]
  \centering
  \includegraphics[width=\textwidth]{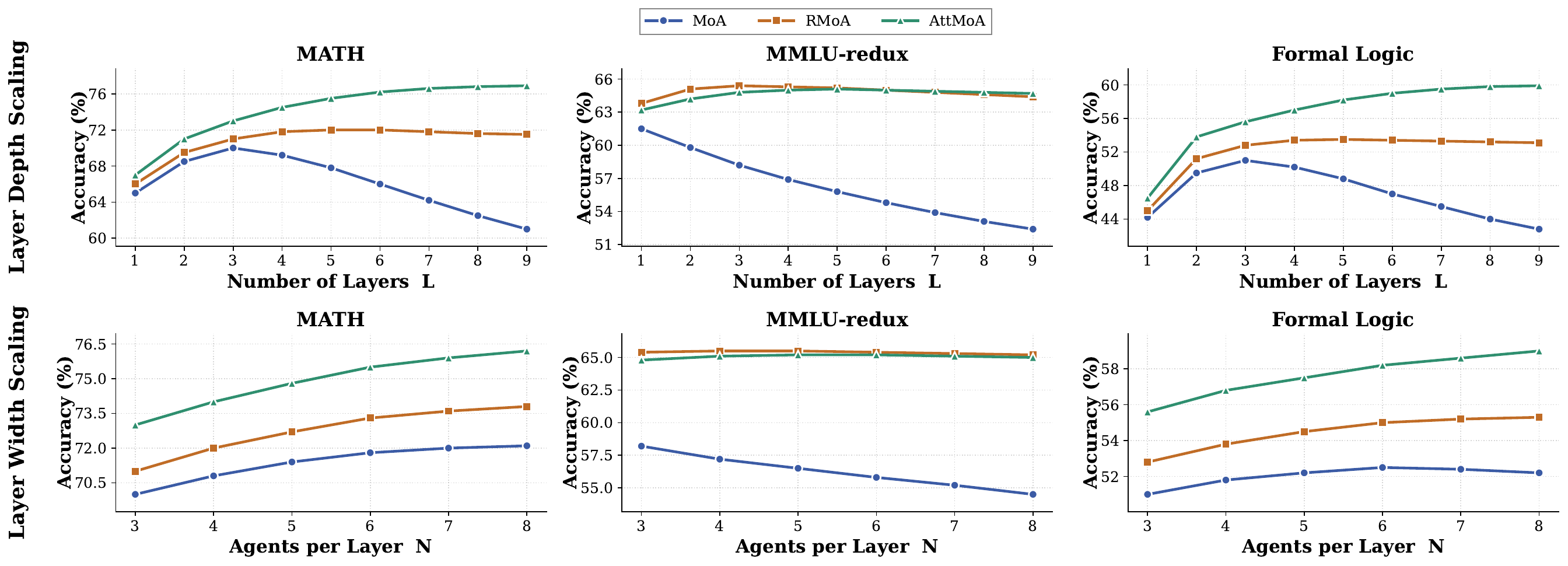}
  \caption{Scaling behavior of three existing MoA variants on 
  three benchmarks (MATH, MMLU-redux, Formal Logic) along two 
  scaling axes. All variants share the same heterogeneous 
  proposer pool: Qwen-2.5-7B, Qwen-2.5-Coder-7B, 
  and Llama-3.1-8B. \textbf{Top row}: layer depth 
  scaling ($L$ from 1 to 9, width fixed at $N=3$). 
  \textbf{Bottom row}: layer width scaling ($N$ from 3 to 8, 
  depth fixed at $L=3$). 
  The depth axis exposes three failure 
  patterns: Standard MoA peaks at shallow depths and then 
  degrades, RMoA plateaus early, and AttentionMoA improves but saturates with limited gains at depth. The width axis shows the same 
  hierarchy at smaller magnitude.
  }
  \label{fig:scaling_limit}
\end{figure*}

\section{Preliminaries: MoA Scaling}
\label{sec:background}

\paragraph{The standard MoA framework.}
The original Mixture-of-Agents framework~\citep{wang2025mixture} 
defines the \emph{standard} MoA structure: $N$ proposer agents 
are organized into $L$ sequential layers followed by a final 
aggregator. Layer 1 receives the original problem $x$ and 
produces $N$ candidate responses; each subsequent layer 
$l \in \{2, \ldots, L\}$ receives the previous layer's outputs 
together with $x$ and produces $N$ new responses; the aggregator 
then synthesizes the outputs of layer $L$ into a final answer. 
Specific variants modify aspects of this structure 
(Section~\ref{sec:related}).

\paragraph{Quality and diversity drive scaling.}
\citet{li2025rethinking} model MoA accuracy through a linear 
regression
\begin{equation}
t = \alpha q + \beta d + \gamma,
\label{eq:moa-perf}
\end{equation}
where $q$ is the average proposer accuracy and $d$ is the Vendi 
Score of proposer outputs. With $\alpha > \beta$ across datasets, 
MoA performance depends on both factors but is more sensitive to 
quality. Sustained scaling therefore requires mechanisms that 
preserve quality and diversity simultaneously, since gains in one 
cannot offset losses in the other.

\paragraph{Empirical scaling failures.}
Figure~\ref{fig:scaling_limit} plots the accuracy of three 
representative MoA variants under two natural scaling axes: 
layer depth (top row, $L$ from 1 to 9 with width fixed at $N=3$) 
and layer width (bottom row, $N$ from 3 to 8 with depth fixed at 
$L=3$). The depth axis exposes three distinct failure patterns. 
Standard MoA~\citep{wang2025mixture} peaks at shallow depths and 
then degrades, with the drop reaching nearly 10 percentage 
points on MMLU-redux. RMoA~\citep{xie2025rmoa} avoids this 
degradation but plateaus early, because its adjacent-layer 
residual cannot retrieve high-quality reasoning produced in 
non-adjacent earlier layers. AttentionMoA~\citep{wen2026attention} 
improves but saturates before deeper layers (with slight late decline on MMLU-redux), since intra-layer attention promotes convergence among the layer's agents. Width scaling exhibits the same hierarchy at smaller 
magnitude: Standard MoA still degrades on MMLU-redux, while RMoA 
and AttentionMoA improve more slowly with diminishing returns.

\paragraph{Design requirements.}
The three patterns above instantiate the failure mechanisms 
outlined in Section~\ref{sec:intro} and show that existing 
variants violate the quality-diversity balance of 
Equation~\eqref{eq:moa-perf} as MoA scales. Sustaining scaling 
therefore requires preserving both factors of this equation:

\textbf{Q (Quality Preservation):} A scalable MoA should retain 
the quality of reasoning produced at each layer so that deeper 
layers build on it rather than dilute it. Both high-quality successful traces (positive guidance) and low-quality failed traces (counter-examples) enrich the input signal beyond positive examples alone. This 
calls for presenting traces in a quality-ranked form rather than 
raw concatenation, and keeping them accessible across all earlier 
layers.

\textbf{D (Diversity Preservation):} A scalable MoA should also 
preserve exploration diversity among agents within each layer, so 
that they do not collapse onto a single solution path as memory 
accumulates. To this end, different agents in the same layer 
should engage with different selections from the memory rather 
than sharing identical inputs.

ReM-MoA, introduced in Section~\ref{sec:method}, addresses 
both Q and D through a Ranked Reasoning Memory paired with a 
curated diversified memory routing scheme: the memory ranks 
and retains traces from every layer, and the routing scheme 
delivers quality-aware reference sets to within-layer 
agents.

\begin{figure*}[t]
  \centering
  \includegraphics[width=\textwidth]{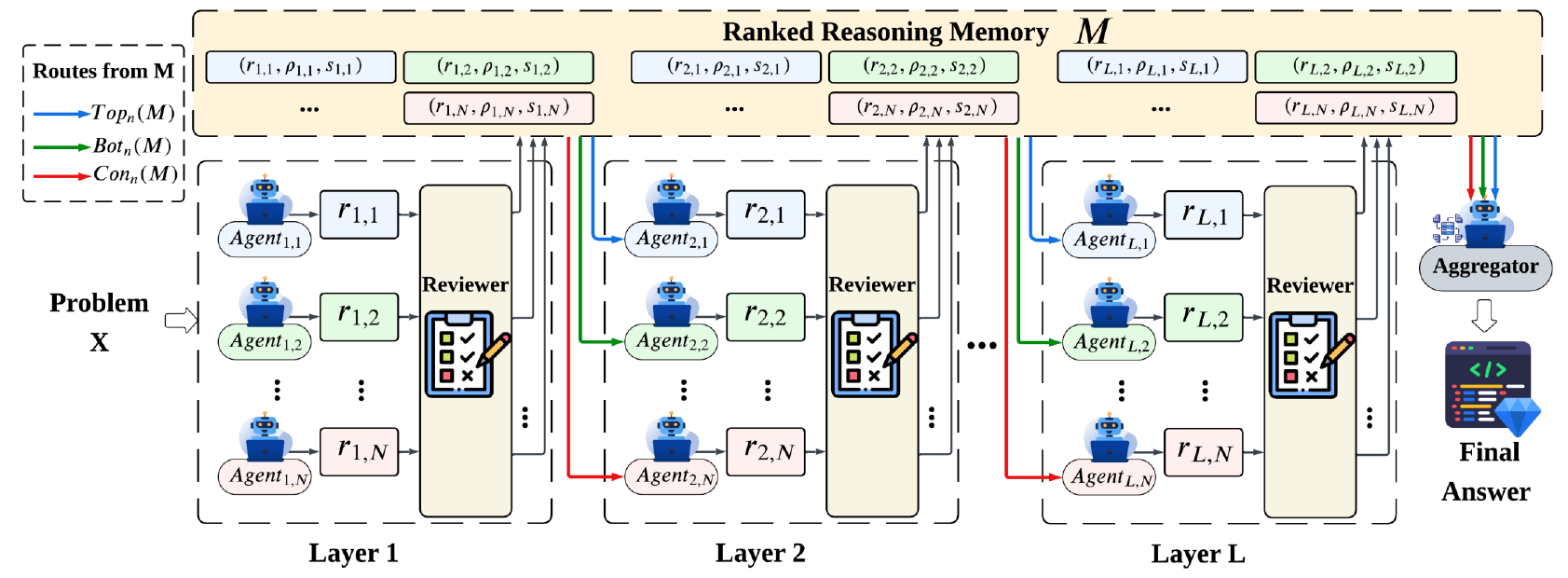}
\caption{The ReM-MoA framework. At each layer, $N$ 
proposer agents produce reasoning traces $r_{l,j}$, and the 
Reviewer Agent compares them and assigns per-trace 
scores $s_{l,j}$ and rationales $\rho_{l,j}$, committed to 
the Ranked Reasoning Memory $\mathcal{M}$. From layer~$2$ 
onwards, each proposer agent receives a curated reference 
set from $\mathcal{M}$ drawn from three families: 
$\text{Top}_n$ (highest-scoring traces), $\text{Bot}_n$ 
(lowest-scoring), or $\text{Con}_n$ (contrastive, 
$\text{Top}_a\cup\text{Bot}_b$). 
}
  \label{fig:framework}
\end{figure*}

\section{ReM-MoA}
\label{sec:method}

ReM-MoA addresses the failure patterns observed in 
Section~\ref{sec:background} through three interconnected 
components: a Ranked Reasoning Memory $\mathcal{M}$, a 
curated diversified memory routing scheme $\Pi$, and an 
optional reviewer distillation pipeline. Let 
$\mathcal{A}_l = \{A_{l,1}, \ldots, A_{l,N}\}$ denote the $N$ 
proposer agents at layer $l \in \{1, \ldots, L\}$, and let 
$r_{l,j} = A_{l,j}(P_{l,j})$ denote the reasoning trace 
produced by agent $A_{l,j}$ given prompt $P_{l,j}$. We write 
$\mathbf{r}_l = (r_{l,1}, \ldots, r_{l,N})$ for the vector of 
all traces produced at layer $l$. Figure~\ref{fig:framework} 
illustrates the framework: the memory ranks and retains 
$\mathbf{r}_l$ at every layer (Section~\ref{sec:memory}); the 
routing scheme delivers distinct curated reference sets to 
within-layer agents (Section~\ref{sec:routing}); and the 
distillation pipeline optionally enhances the memory's 
ranking module (Section~\ref{sec:distillation}). After 
layer $L$, the aggregator consults the memory $\mathcal{M}$ 
through routes drawn from the same three families as the 
proposer agents, and synthesizes the final answer.

\subsection{Ranked Reasoning Memory}
\label{sec:memory}

\paragraph{Memory subsystem.}
The Ranked Reasoning Memory $\mathcal{M}$ is a shared 
subsystem that records ranked traces at every proposer layer. 
After layer $l$ has been processed, the memory accumulates as 
the union of per-layer entries:
\begin{equation}
\mathcal{M}_{\leq l} = \bigcup_{k=1}^{l} \mathcal{M}_k,
\label{eq:memory-set}
\end{equation}
where $\mathcal{M}_k$ is the entry committed by layer $k$ 
(defined below). The memory is append-only and supports read 
access from any subsequent layer: for any $l' > l$, layer $l'$ 
may consult $\mathcal{M}_{\leq l}$. This subsystem addresses Q (Quality Preservation) 
because traces are stored in quality-ranked form rather than 
raw concatenated outputs, and remain accessible across 
arbitrary layer distances rather than only between adjacent 
layers.

\paragraph{The Reviewer Agent.}
The ranking inside the memory is performed by an LLM-based 
Reviewer Agent $\mathcal{R}$. After the $N$ proposers at 
layer $l$ produce their traces $\mathbf{r}_l$, the Reviewer 
is invoked once with all $N$ traces presented together; it 
compares the traces against each other and emits, for each 
trace, an integer score together with a targeted rationale 
specific to that trace:
\begin{equation}
(\mathbf{s}_l, \boldsymbol{\rho}_l) = \mathcal{R}(x, \mathbf{r}_l),
\label{eq:reviewer}
\end{equation}
where $\mathbf{s}_l = (s_{l,1}, \ldots, s_{l,N})$ with each 
$s_{l,j} \in \{1, \ldots, S\}$ being an integer score on a 
scale with maximum $S$ (higher is better), and 
$\boldsymbol{\rho}_l = (\rho_{l,1}, \ldots, \rho_{l,N})$ with 
each $\rho_{l,j}$ being a rationale specific to trace 
$r_{l,j}$ (rationales are independent across traces). We use 
\emph{comparative} scoring rather than independent scoring 
because grounding each trace's score in relative differences 
across candidates produces more reliable estimates than 
absolute scoring~\citep{zheng2023judging}.

\paragraph{Memory entries and access.}
Each layer commits one entry to the memory, formed by zipping 
the traces with their Reviewer-assigned scores and rationales:
\begin{equation}
\mathcal{M}_l = \{(r_{l,j}, s_{l,j}, \rho_{l,j})\}_{j=1}^N.
\label{eq:memory-entry}
\end{equation}
$\mathcal{M}_l$ is a set of $N$ trace-score-rationale triples. 
The accumulated memory $\mathcal{M}_{\leq l}$ thus contains 
$N \cdot l$ triples available for cross-layer retrieval by 
the routing scheme described next.

\subsection{Curated Diversified Memory Routing}
\label{sec:routing}

\paragraph{Curated reference delivery.}
At every subsequent layer, each proposer agent receives, in 
addition to the problem $x$, a curated set of reference 
traces drawn from the memory. The selection follows two 
design principles. First, the reference set must carry 
quality-aware information for the agent's reasoning: it 
comprises high-scoring traces as positive examples, 
low-scoring traces as negative examples, or both. Second, 
the reference sets must differ across the $N$ agents within 
the same layer, so the layer as a whole explores the 
memory's content rather than redundantly consuming the same 
references. Formally, at layer $l \geq 2$, the curated 
diversified memory routing scheme $\Pi$ operates on the 
\emph{global pool} of all previously committed entries, 
$\mathcal{P}_l = \mathcal{M}_{\leq l-1}$:
\begin{equation}
\mathcal{S}_{l, j} = \Pi_j(\mathcal{P}_l),
\label{eq:routing}
\end{equation}
where $\mathcal{S}_{l, j} \subset \mathcal{P}_l$ is the 
curated reference set routed to agent $A_{l, j}$. Operating 
on $\mathcal{P}_l$ rather than only the most recent layer 
realizes the cross-layer access promised by the memory 
subsystem (Section~\ref{sec:memory}). The prompt for agent 
$A_{l, j}$ is then:
\begin{equation}
P_{l, j} = 
\begin{cases}
x & \text{if } l = 1, \\
x \oplus \mathcal{S}_{l, j} & \text{if } l \geq 2,
\end{cases}
\label{eq:prompt}
\end{equation}
where $\oplus$ denotes concatenation; each entry in 
$\mathcal{S}_{l, j}$ contributes its trace together with its 
score and rationale.

\paragraph{Routing functions.}
We implement the two design principles above through three 
families of routing functions operating on $\mathcal{P}_l$. 
Let $\text{Top}_n(\mathcal{M})$ denote the size-$n$ subset 
of $\mathcal{M}$ containing the highest-scoring entries, and 
$\text{Bot}_n(\mathcal{M})$ the analogous subset for the 
lowest-scoring ones. The three families are: 
\emph{contrastive}, of the form 
$\text{Top}_a \cup \text{Bot}_b$ (denoted 
$\text{Con}_n$ with $n=a+b$), pairing high- and low-scoring 
entries; \emph{positive reinforcement}, of the form 
$\text{Top}_a$, emphasizing successful patterns; and 
\emph{failure-mode focus}, of the form $\text{Bot}_a$, 
surfacing common pitfalls. For each layer, the $N$ routing 
functions $\Pi_1, \ldots, \Pi_N$ are distributed across 
these three families, with the distribution chosen to 
balance the three modes of reference. As one concrete 
instance, for $N=3$ we use one route per family:
\begin{equation}
\Pi_j(\mathcal{P}_l) = 
\begin{cases}
\text{Top}_1(\mathcal{P}_l) \cup \text{Bot}_1(\mathcal{P}_l) & j = 1, \\
\text{Top}_2(\mathcal{P}_l) & j = 2, \\
\text{Bot}_2(\mathcal{P}_l) & j = 3.
\end{cases}
\label{eq:routes-N3}
\end{equation}
Appendix~\ref{app:routing} details the distributions used 
for larger $N$ in our scaling experiments. This design 
realizes both Q (Quality Preservation) and D (Diversity 
Preservation). For Q, the curated reference sets preserve 
the Reviewer-extracted quality signal as it propagates 
forward through the layers. For D, the $N$ distinct routes 
within each layer sustain exploration diversity even as the 
memory grows.

\subsection{Reviewer Distillation (Optional Enhancement)}
\label{sec:distillation}

\paragraph{Distillation pipeline.}
The Reviewer Agent $\mathcal{R}$ is implemented as an 
off-the-shelf LLM by default. It can be optionally enhanced 
via frontier-model distillation~\citep{hinton2015distilling}. 
Crucially, the training data for distillation comes from 
\emph{ReM-MoA's own runs}: we execute the framework on a 
multi-domain set of training problems, collect the proposer 
traces $\mathbf{r}_l$ generated at each layer, and query a 
strong external teacher $\mathcal{T}$ on each (problem, traces) 
pair to obtain high-quality supervision targets:
\begin{equation}
\mathcal{D}_{\text{train}} = \big\{(x_i, \mathbf{r}_i, 
\mathbf{s}_i^*, \boldsymbol{\rho}_i^*) \mid 
(\mathbf{s}_i^*, \boldsymbol{\rho}_i^*) = \mathcal{T}(x_i, 
\mathbf{r}_i)\big\},
\label{eq:train-data}
\end{equation}
where $(x_i, \mathbf{r}_i)$ is sampled from the framework's 
intermediate outputs and 
$(\mathbf{s}_i^*, \boldsymbol{\rho}_i^*)$ is the teacher's 
score-rationale labeling treated as the supervision target. 
A smaller base Reviewer $\mathcal{R}_\theta$ is then fine-tuned 
via parameter-efficient LoRA~\citep{hu2022lora} to imitate the 
teacher:
\begin{equation}
\theta^* = \arg\min_{\theta} \sum_{(x, \mathbf{r}, 
\mathbf{s}^*, \boldsymbol{\rho}^*) \in \mathcal{D}_{\text{train}}} 
\mathcal{L}\big(\mathcal{R}_\theta(x, \mathbf{r}),\ 
(\mathbf{s}^*, \boldsymbol{\rho}^*)\big),
\label{eq:lora-loss}
\end{equation}
where $\mathcal{L}$ is the standard supervised loss on the 
concatenated score-rationale output. Because the training 
distribution matches the traces ReM-MoA actually produces at 
inference, the resulting $\mathcal{R}_{\theta^*}$ learns to 
judge exactly the distribution of traces it will encounter at 
deployment.

\paragraph{Multi-domain training.}
The training problems span multiple reasoning domains. Multi-domain training is essential to the design: it lets a single Reviewer generalize across all benchmarks we evaluate, including held-out domains not seen during training (Section~\ref{sec:experiments}). Full training configuration and 
hyperparameters are described in Section~\ref{sec:setup}.

\begin{table}[t]
\centering
\caption{Depth scaling ($N{=}3$ fixed). Trend: $\downarrow$ degrade, $\rightarrow$ plateau, $\uparrow\rightarrow$ saturate, $\uparrow\uparrow$ sustain (based on tail-3 gain). Our methods in \textbf{bold}; $^*$: with distilled Reviewer.}
\label{tab:depth-scaling}
\setlength{\tabcolsep}{2pt}
\renewcommand{\arraystretch}{1.1}
\scriptsize
\resizebox{\columnwidth}{!}{%
\begin{tabular}{ll|ccccccccc|c}
\toprule
\textbf{Benchmark} & \textbf{Method} & $L{=}1$ & $L{=}2$ & $L{=}3$ & $L{=}4$ & $L{=}5$ & $L{=}6$ & $L{=}7$ & $L{=}8$ & $L{=}9$ & \textbf{Trend} \\
\midrule
\multirow{5}{*}{\textit{MATH}} & Standard MoA & 65.0 & 68.5 & 70.0 & 69.2 & 67.8 & 66.0 & 64.2 & 62.5 & 61.0 & $\downarrow$ \\
 & RMoA & 66.0 & 69.5 & 71.0 & 71.8 & 72.0 & 72.0 & 71.8 & 71.6 & 71.5 & $\uparrow\rightarrow$ \\
 & AttentionMoA & 67.0 & 71.0 & 73.0 & 74.5 & 75.5 & 76.2 & 76.6 & 76.8 & 76.9 & $\uparrow\rightarrow$ \\
 & \textbf{ReM-MoA} & \textbf{68.0} & \textbf{72.5} & \textbf{75.0} & \textbf{77.0} & \textbf{78.5} & \textbf{79.5} & \textbf{80.2} & \textbf{80.7} & \textbf{81.0} & \textbf{$\uparrow\uparrow$} \\
 & \textbf{ReM-MoA$^*$} & \textbf{68.3} & \textbf{73.5} & \textbf{76.8} & \textbf{79.2} & \textbf{81.0} & \textbf{82.2} & \textbf{83.0} & \textbf{83.7} & \textbf{84.0} & \textbf{$\uparrow\uparrow$} \\
\midrule
\multirow{5}{*}{\textit{MMLU-redux}} & Standard MoA & 61.5 & 59.8 & 58.2 & 56.9 & 55.8 & 54.8 & 53.9 & 53.1 & 52.4 & $\downarrow$ \\
 & RMoA & 63.8 & 65.1 & 65.4 & 65.3 & 65.2 & 65.0 & 64.8 & 64.6 & 64.4 & $\rightarrow$ \\
 & AttentionMoA & 63.2 & 64.2 & 64.8 & 65.0 & 65.1 & 65.0 & 64.9 & 64.8 & 64.7 & $\rightarrow$ \\
 & \textbf{ReM-MoA} & \textbf{64.0} & \textbf{66.0} & \textbf{67.0} & \textbf{67.8} & \textbf{68.3} & \textbf{68.6} & \textbf{68.8} & \textbf{68.9} & \textbf{68.9} & \textbf{$\uparrow\rightarrow$} \\
 & \textbf{ReM-MoA$^*$} & \textbf{64.2} & \textbf{66.3} & \textbf{67.4} & \textbf{68.3} & \textbf{68.9} & \textbf{69.3} & \textbf{69.6} & \textbf{69.9} & \textbf{70.2} & \textbf{$\uparrow\uparrow$} \\
\midrule
\multirow{5}{*}{\textit{Formal Logic}} & Standard MoA & 44.2 & 49.5 & 51.0 & 50.2 & 48.8 & 47.0 & 45.5 & 44.0 & 42.8 & $\downarrow$ \\
 & RMoA & 45.0 & 51.2 & 52.8 & 53.4 & 53.5 & 53.4 & 53.3 & 53.2 & 53.1 & $\uparrow\rightarrow$ \\
 & AttentionMoA & 46.5 & 53.8 & 55.6 & 57.0 & 58.2 & 59.0 & 59.5 & 59.8 & 59.9 & $\uparrow\rightarrow$ \\
 & \textbf{ReM-MoA} & \textbf{47.8} & \textbf{55.5} & \textbf{59.0} & \textbf{61.5} & \textbf{63.5} & \textbf{65.0} & \textbf{66.0} & \textbf{66.8} & \textbf{67.2} & \textbf{$\uparrow\uparrow$} \\
 & \textbf{ReM-MoA$^*$} & \textbf{48.2} & \textbf{56.7} & \textbf{60.8} & \textbf{63.7} & \textbf{66.0} & \textbf{67.7} & \textbf{68.8} & \textbf{69.7} & \textbf{70.2} & \textbf{$\uparrow\uparrow$} \\
\midrule
\multirow{5}{*}{\textit{CRUX}} & Standard MoA & 54.0 & 56.5 & 57.5 & 57.0 & 56.0 & 54.8 & 53.5 & 52.2 & 51.0 & $\downarrow$ \\
 & RMoA & 54.5 & 57.5 & 58.5 & 59.0 & 59.2 & 59.4 & 59.4 & 59.5 & 59.5 & $\uparrow\rightarrow$ \\
 & AttentionMoA & 55.5 & 58.8 & 60.5 & 61.8 & 62.7 & 63.3 & 63.6 & 63.8 & 63.9 & $\uparrow\rightarrow$ \\
 & \textbf{ReM-MoA} & \textbf{56.5} & \textbf{60.5} & \textbf{63.0} & \textbf{65.2} & \textbf{67.0} & \textbf{68.5} & \textbf{69.5} & \textbf{70.0} & \textbf{70.2} & \textbf{$\uparrow\uparrow$} \\
 & \textbf{ReM-MoA$^*$} & \textbf{57.0} & \textbf{61.6} & \textbf{64.6} & \textbf{67.2} & \textbf{69.3} & \textbf{71.1} & \textbf{72.2} & \textbf{72.8} & \textbf{73.0} & \textbf{$\uparrow\uparrow$} \\
\midrule
\multirow{5}{*}{\textit{HellaSwag}} & Standard MoA & 73.5 & 76.0 & 76.8 & 76.2 & 75.3 & 74.2 & 73.0 & 71.8 & 70.8 & $\downarrow$ \\
 & RMoA & 74.0 & 76.8 & 77.5 & 77.8 & 78.0 & 77.9 & 77.8 & 77.7 & 77.6 & $\uparrow\rightarrow$ \\
 & AttentionMoA & 74.8 & 77.5 & 78.5 & 79.2 & 79.7 & 80.0 & 80.1 & 80.2 & 80.2 & $\uparrow\rightarrow$ \\
 & \textbf{ReM-MoA} & \textbf{75.5} & \textbf{78.8} & \textbf{80.5} & \textbf{81.5} & \textbf{82.2} & \textbf{82.6} & \textbf{82.8} & \textbf{82.9} & \textbf{82.9} & \textbf{$\uparrow\rightarrow$} \\
 & \textbf{ReM-MoA$^*$} & \textbf{75.8} & \textbf{79.5} & \textbf{81.2} & \textbf{82.2} & \textbf{83.0} & \textbf{83.4} & \textbf{83.7} & \textbf{84.0} & \textbf{84.3} & \textbf{$\uparrow\uparrow$} \\
\bottomrule
\end{tabular}
}
\end{table}

\begin{table}[t]
\centering
\caption{Width scaling ($L{=}3$ fixed). Conventions follow Table~\ref{tab:depth-scaling}.}
\label{tab:width-scaling}
\setlength{\tabcolsep}{3pt}
\renewcommand{\arraystretch}{1.1}
\scriptsize
\resizebox{\columnwidth}{!}{%
\begin{tabular}{ll|cccccc|c}
\toprule
\textbf{Benchmark} & \textbf{Method} & $N{=}3$ & $N{=}4$ & $N{=}5$ & $N{=}6$ & $N{=}7$ & $N{=}8$ & \textbf{Trend} \\
\midrule
\multirow{5}{*}{\textit{MATH}} & Standard MoA & 70.0 & 70.8 & 71.4 & 71.8 & 72.0 & 72.1 & $\uparrow\rightarrow$ \\
 & RMoA & 71.0 & 72.0 & 72.7 & 73.3 & 73.6 & 73.8 & $\uparrow\uparrow$ \\
 & AttentionMoA & 73.0 & 74.0 & 74.8 & 75.5 & 75.9 & 76.2 & $\uparrow\uparrow$ \\
 & \textbf{ReM-MoA} & \textbf{75.0} & \textbf{76.2} & \textbf{77.2} & \textbf{78.0} & \textbf{78.5} & \textbf{78.9} & \textbf{$\uparrow\uparrow$} \\
 & \textbf{ReM-MoA$^*$} & \textbf{76.8} & \textbf{78.0} & \textbf{79.1} & \textbf{79.9} & \textbf{80.5} & \textbf{80.9} & \textbf{$\uparrow\uparrow$} \\
\midrule
\multirow{5}{*}{\textit{MMLU-redux}} & Standard MoA & 58.2 & 57.2 & 56.5 & 55.8 & 55.2 & 54.5 & $\downarrow$ \\
 & RMoA & 65.4 & 65.5 & 65.5 & 65.4 & 65.3 & 65.2 & $\rightarrow$ \\
 & AttentionMoA & 64.8 & 65.1 & 65.2 & 65.2 & 65.1 & 65.0 & $\rightarrow$ \\
 & \textbf{ReM-MoA} & \textbf{67.0} & \textbf{67.8} & \textbf{68.3} & \textbf{68.8} & \textbf{69.0} & \textbf{69.2} & \textbf{$\uparrow\rightarrow$} \\
 & \textbf{ReM-MoA$^*$} & \textbf{67.4} & \textbf{68.2} & \textbf{68.7} & \textbf{69.3} & \textbf{69.5} & \textbf{69.8} & \textbf{$\uparrow\uparrow$} \\
\midrule
\multirow{5}{*}{\textit{Formal Logic}} & Standard MoA & 51.0 & 51.8 & 52.2 & 52.5 & 52.4 & 52.2 & $\rightarrow$ \\
 & RMoA & 52.8 & 53.8 & 54.5 & 55.0 & 55.2 & 55.3 & $\uparrow\rightarrow$ \\
 & AttentionMoA & 55.6 & 56.8 & 57.5 & 58.2 & 58.6 & 59.0 & $\uparrow\uparrow$ \\
 & \textbf{ReM-MoA} & \textbf{59.0} & \textbf{60.5} & \textbf{61.8} & \textbf{63.0} & \textbf{63.8} & \textbf{64.5} & \textbf{$\uparrow\uparrow$} \\
 & \textbf{ReM-MoA$^*$} & \textbf{60.8} & \textbf{62.3} & \textbf{63.7} & \textbf{64.9} & \textbf{65.8} & \textbf{66.6} & \textbf{$\uparrow\uparrow$} \\
\midrule
\multirow{5}{*}{\textit{CRUX}} & Standard MoA & 57.5 & 58.0 & 58.4 & 58.6 & 58.5 & 58.3 & $\rightarrow$ \\
 & RMoA & 58.5 & 59.3 & 59.8 & 60.3 & 60.6 & 60.7 & $\uparrow\rightarrow$ \\
 & AttentionMoA & 60.5 & 61.5 & 62.3 & 62.9 & 63.3 & 63.6 & $\uparrow\uparrow$ \\
 & \textbf{ReM-MoA} & \textbf{63.0} & \textbf{64.3} & \textbf{65.3} & \textbf{66.2} & \textbf{66.8} & \textbf{67.3} & \textbf{$\uparrow\uparrow$} \\
 & \textbf{ReM-MoA$^*$} & \textbf{64.6} & \textbf{66.0} & \textbf{67.0} & \textbf{68.0} & \textbf{68.7} & \textbf{69.3} & \textbf{$\uparrow\uparrow$} \\
\midrule
\multirow{5}{*}{\textit{HellaSwag}} & Standard MoA & 76.8 & 77.2 & 77.4 & 77.5 & 77.4 & 77.2 & $\rightarrow$ \\
 & RMoA & 77.5 & 78.0 & 78.4 & 78.7 & 78.8 & 78.9 & $\rightarrow$ \\
 & AttentionMoA & 78.5 & 79.2 & 79.7 & 80.1 & 80.3 & 80.4 & $\uparrow\rightarrow$ \\
 & \textbf{ReM-MoA} & \textbf{80.5} & \textbf{81.2} & \textbf{81.8} & \textbf{82.3} & \textbf{82.6} & \textbf{82.9} & \textbf{$\uparrow\uparrow$} \\
 & \textbf{ReM-MoA$^*$} & \textbf{81.2} & \textbf{82.0} & \textbf{82.6} & \textbf{83.1} & \textbf{83.4} & \textbf{83.7} & \textbf{$\uparrow\uparrow$} \\
\bottomrule
\end{tabular}
}
\end{table}

\section{Experiments}
\label{sec:experiments}

We evaluate ReM-MoA against existing MoA variants to answer 
three research questions:

\paragraph{RQ1.} Does ReM-MoA continue to improve as depth and 
width scale, where prior MoA variants degrade, plateau, or 
saturate?

\paragraph{RQ2.} How much do the Ranked Reasoning Memory and 
the Curated Diversified Memory Routing each contribute to 
ReM-MoA's overall gain?

\paragraph{RQ3.} Does multi-domain Reviewer distillation 
transfer to held-out reasoning domains?

\begin{figure*}[t]
  \centering
  \includegraphics[width=\textwidth]{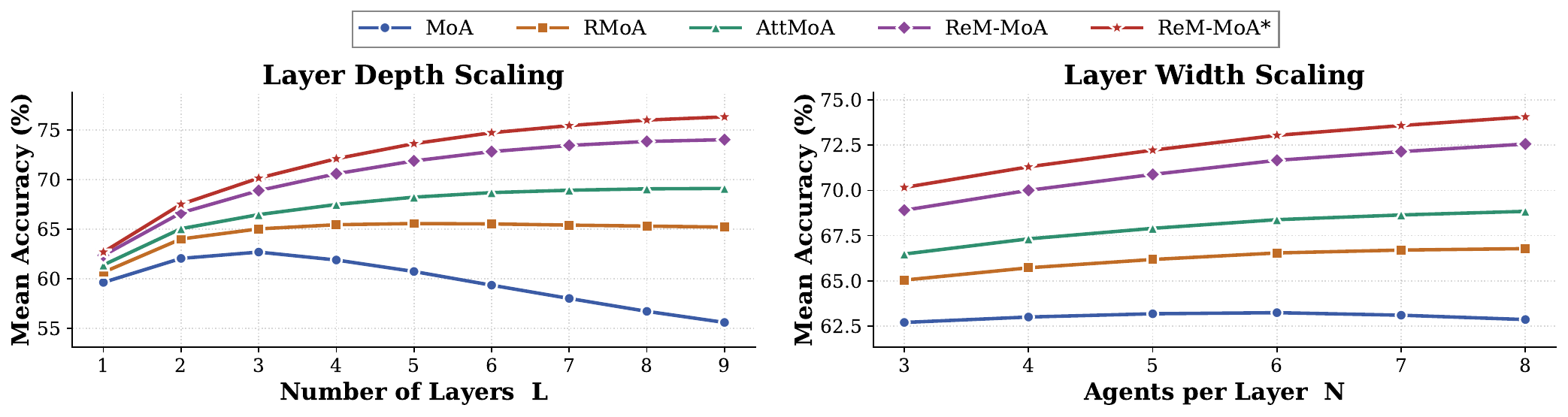}
  \caption{Mean accuracy across the five benchmarks under depth 
  scaling (left, $N{=}3$ fixed) and width scaling (right, 
  $L{=}3$ fixed). Mean accuracy under depth scaling exhibits a clear separation: Standard MoA peaks at $L{=}3$ and 
  declines, RMoA and AttentionMoA flatten as $L$ grows, while 
  ReM-MoA and ReM-MoA$^*$ continue to improve.}
  \label{fig:mean-scaling}
\end{figure*}

\subsection{Experimental Setup}
\label{sec:setup}

\paragraph{Benchmarks.}
We evaluate on five reasoning benchmarks, one per target domain: 
MATH~\citep{hendrycks2021measuring} (mathematics), 
MMLU-redux~\citep{gema2025done} (knowledge), the 
\texttt{formal\_logic} subject of MMLU~\citep{hendrycks2020measuring} 
(formal logic), CRUX~\citep{gu2024cruxeval} (code understanding), 
and HellaSwag~\citep{zellers2019hellaswag} (common-sense reasoning). We report accuracy on all evaluation domains, except for the code 
benchmarks (CRUX and, in the distillation experiment, MBPP), for 
which we report pass@1.

\paragraph{Proposer pool and aggregator.}
A heterogeneous pool of three 7-8B open-source instruction-tuned 
models (Qwen2.5-7B-Instruct~\citep{yang2024qwen2}, 
Llama-3.1-8B-Instruct~\citep{grattafiori2024llama}, and 
Qwen2.5-Coder-7B-Instruct~\citep{hui2024qwen}) serves as proposers. 
Qwen2.5-7B-Instruct is the aggregator in all configurations.

\paragraph{Reviewer Agent.}
We evaluate two variants of our framework that differ only in the 
Reviewer: \textbf{ReM-MoA} uses the off-the-shelf 
Qwen2.5-7B-Instruct as the Reviewer, while \textbf{ReM-MoA$^*$} 
replaces it with a LoRA-distilled fine-tune of the same base 
model. For the distillation, we query GPT-5.5 as the teacher 
$\mathcal{T}$ on traces generated by ReM-MoA itself, then 
fine-tune Qwen2.5-7B-Instruct with LoRA~\citep{hu2022lora}. 
Training data consists of 100 problems each from MATH, 
ARC-Challenge~\citep{clark2018think}, MedQA~\citep{jin2021disease}, 
MBPP~\citep{austin2021program}, and HellaSwag (500 problems total); 
ReM-MoA is run on each at $L=3, N=3$, yielding $1{,}500$ 
(problem, layer-traces) training tuples. Full LoRA configuration 
is in Appendix~\ref{app:setup}.

\paragraph{Baselines.}
We compare against three MoA variants representing the paradigms 
introduced in Section~\ref{sec:related}: standard 
MoA~\citep{wang2025mixture} (information concatenation), 
RMoA~\citep{xie2025rmoa} (information consolidation), and 
AttentionMoA~\citep{wen2026attention} (inter-agent interaction). 
All baselines share our proposer pool and aggregator, retaining 
each baseline's original hyperparameters.

\paragraph{Implementation.}
All experiments run on NVIDIA A100 GPUs via 
vLLM~\citep{kwon2023efficient}. Proposer agents sample at 
temperature 0.8, the Reviewer at 0.3, and the aggregator at 0.5. 
We evaluate along two axes: \emph{depth}, sweeping 
$L \in \{1, \ldots, 9\}$ at $N=3$, and \emph{width}, sweeping 
$N \in \{3, \ldots, 8\}$ at $L=3$. Further implementation details 
are in Appendix~\ref{app:setup}.

\subsection{Sustained Scaling and Cross-Benchmark Results (RQ1)}
\label{sec:scaling}

Tables~\ref{tab:depth-scaling} and~\ref{tab:width-scaling} 
report depth and width scaling across the five benchmarks. 
Figure~\ref{fig:mean-scaling} plots the corresponding mean 
accuracy.

\paragraph{Baseline failure modes are confirmed on all five 
benchmarks.}
Standard MoA degrades on all five benchmarks, peaking by 
$L{=}3$ and losing $9.1$ points on MMLU-redux ($61.5 \to 
52.4$), $4.0$ on MATH, $3.0$ on CRUX, $2.7$ on HellaSwag, 
and $1.4$ on Formal Logic by $L{=}9$. RMoA plateaus on 
MMLU-redux (its representative failure mode in 
Figure~\ref{fig:scaling_limit}) and saturates on the other 
four benchmarks, with tail-3 gain below $+0.2$ on every 
non-plateau cell. AttentionMoA saturates on four benchmarks 
and plateaus on MMLU-redux (peaking at $L{=}5$ then slightly 
declining), with per-step gain shrinking to $\leq 0.1$ at 
$L{=}9$, indicating intra-layer convergence among agents.

\paragraph{ReM-MoA outperforms every baseline across all 
configurations.}
ReM-MoA$^*$ is column-best with ReM-MoA second across all 
$45$ benchmark-$L$ and $30$ benchmark-$N$ cells. The lead is 
largest on reasoning-intensive benchmarks: at $L{=}9$, 
ReM-MoA exceeds AttentionMoA by $+7.3$ on Formal Logic 
($67.2$ vs $59.9$), $+6.3$ on CRUX, and $+4.1$ on MATH. On 
knowledge-heavy MMLU-redux and common-sense HellaSwag, where 
headroom is smaller, ReM-MoA still leads at every $L$ but 
saturates due to limited remaining ceiling.

\paragraph{The gap widens monotonically with scale.}
Figure~\ref{fig:mean-scaling} renders the spectrum of scaling behaviors in one view and shows a clean monotonic fan-out among the 
five methods along the depth axis. Quantitatively, ReM-MoA's 
gap over the strongest baseline grows from $+0.84$ at $L{=}1$ 
to $+4.92$ at $L{=}9$ ($5.9{\times}$); against Standard MoA 
it expands from $+2.72$ to $+18.44$ ($6.8{\times}$). Width 
yields a more modest expansion from $+2.30$ at $N{=}3$ to 
$+3.68$ at $N{=}8$, with the five methods following more 
closely-spaced trajectories than under depth (right panel of 
Fig.~\ref{fig:mean-scaling}). The widening holds against all 
three baselines across the five benchmarks.

\paragraph{Distillation extends sustained scaling to all 
benchmarks.}
ReM-MoA$^*$ improves accuracy across all configurations, with 
gains over ReM-MoA of up to $3.0$ points at $L{=}9$ on MATH 
and Formal Logic. Distillation also flips the Trend label on 
the three cells where ReM-MoA still saturated (MMLU-redux 
and HellaSwag depth, MMLU-redux width), all moving to 
$\uparrow\uparrow$. ReM-MoA$^*$ thus sustains scaling on all 
five benchmarks under both axes.

\begin{table}[t]
\centering
\caption{Ablation study under depth scaling ($N{=}3$). All variants share the off-the-shelf Reviewer (ReM-MoA, not ReM-MoA$^*$). Trend conventions follow Table~\ref{tab:depth-scaling}. The \textbf{ReM-MoA (full)} row repeats Table~\ref{tab:depth-scaling} for reference.}
\label{tab:ablation}
\setlength{\tabcolsep}{2pt}
\renewcommand{\arraystretch}{1.1}
\scriptsize
\resizebox{\columnwidth}{!}{%
\begin{tabular}{ll|ccccccccc|c}
\toprule
\textbf{Benchmark} & \textbf{Variant} & $L{=}1$ & $L{=}2$ & $L{=}3$ & $L{=}4$ & $L{=}5$ & $L{=}6$ & $L{=}7$ & $L{=}8$ & $L{=}9$ & \textbf{Trend} \\
\midrule
\multirow{4}{*}{\textit{MATH}} & \textbf{ReM-MoA (full)} & \textbf{68.0} & \textbf{72.5} & \textbf{75.0} & \textbf{77.0} & \textbf{78.5} & \textbf{79.5} & \textbf{80.2} & \textbf{80.7} & \textbf{81.0} & \textbf{$\uparrow\uparrow$} \\
 & $-$ cross-layer access & 68.0 & 72.5 & 74.1 & 75.5 & 76.0 & 76.4 & 76.5 & 76.3 & 76.2 & $\uparrow\rightarrow$ \\
 & $-$ routing diversification & 68.0 & 72.0 & 74.0 & 75.2 & 76.2 & 77.0 & 77.4 & 77.6 & 77.7 & $\uparrow\rightarrow$ \\
 & $-$ rationale & 67.7 & 71.9 & 74.2 & 75.9 & 77.2 & 78.0 & 78.7 & 79.1 & 79.3 & $\uparrow\uparrow$ \\
\midrule
\multirow{4}{*}{\textit{MMLU-redux}} & \textbf{ReM-MoA (full)} & \textbf{64.0} & \textbf{66.0} & \textbf{67.0} & \textbf{67.8} & \textbf{68.3} & \textbf{68.6} & \textbf{68.8} & \textbf{68.9} & \textbf{68.9} & \textbf{$\uparrow\rightarrow$} \\
 & $-$ cross-layer access & 64.0 & 66.0 & 66.4 & 66.7 & 66.9 & 67.0 & 66.9 & 66.7 & 66.8 & $\uparrow\rightarrow$ \\
 & $-$ routing diversification & 64.0 & 65.8 & 66.5 & 66.9 & 67.2 & 67.4 & 67.5 & 67.6 & 67.6 & $\uparrow\rightarrow$ \\
 & $-$ rationale & 63.9 & 65.8 & 66.5 & 67.2 & 67.5 & 67.8 & 68.0 & 68.0 & 68.0 & $\uparrow\rightarrow$ \\
\midrule
\multirow{4}{*}{\textit{Formal Logic}} & \textbf{ReM-MoA (full)} & \textbf{47.8} & \textbf{55.5} & \textbf{59.0} & \textbf{61.5} & \textbf{63.5} & \textbf{65.0} & \textbf{66.0} & \textbf{66.8} & \textbf{67.2} & \textbf{$\uparrow\uparrow$} \\
 & $-$ cross-layer access & 47.8 & 55.5 & 57.7 & 59.6 & 60.6 & 60.9 & 61.1 & 60.8 & 61.0 & $\uparrow\rightarrow$ \\
 & $-$ routing diversification & 47.8 & 55.0 & 57.8 & 59.8 & 61.5 & 62.8 & 63.5 & 63.8 & 63.9 & $\uparrow\rightarrow$ \\
 & $-$ rationale & 47.4 & 54.8 & 57.9 & 60.1 & 61.8 & 63.1 & 64.1 & 64.8 & 65.1 & $\uparrow\uparrow$ \\
\midrule
\multirow{4}{*}{\textit{CRUX}} & \textbf{ReM-MoA (full)} & \textbf{56.5} & \textbf{60.5} & \textbf{63.0} & \textbf{65.2} & \textbf{67.0} & \textbf{68.5} & \textbf{69.5} & \textbf{70.0} & \textbf{70.2} & \textbf{$\uparrow\uparrow$} \\
 & $-$ cross-layer access & 56.5 & 60.5 & 62.1 & 63.5 & 64.2 & 64.7 & 64.5 & 64.4 & 64.5 & $\uparrow\rightarrow$ \\
 & $-$ routing diversification & 56.5 & 60.0 & 62.0 & 63.5 & 64.8 & 65.8 & 66.5 & 66.8 & 66.9 & $\uparrow\rightarrow$ \\
 & $-$ rationale & 56.1 & 59.8 & 62.1 & 63.9 & 65.4 & 66.5 & 67.3 & 67.8 & 68.0 & $\uparrow\uparrow$ \\
\midrule
\multirow{4}{*}{\textit{HellaSwag}} & \textbf{ReM-MoA (full)} & \textbf{75.5} & \textbf{78.8} & \textbf{80.5} & \textbf{81.5} & \textbf{82.2} & \textbf{82.6} & \textbf{82.8} & \textbf{82.9} & \textbf{82.9} & \textbf{$\uparrow\rightarrow$} \\
 & $-$ cross-layer access & 75.5 & 78.8 & 79.7 & 80.3 & 80.6 & 80.5 & 80.4 & 80.2 & 80.3 & $\uparrow\rightarrow$ \\
 & $-$ routing diversification & 75.5 & 78.5 & 79.7 & 80.4 & 81.0 & 81.3 & 81.5 & 81.6 & 81.6 & $\uparrow\rightarrow$ \\
 & $-$ rationale & 75.3 & 78.5 & 79.8 & 80.8 & 81.5 & 81.8 & 82.0 & 82.2 & 82.2 & $\uparrow\rightarrow$ \\
\bottomrule
\end{tabular}
}
\end{table}

\subsection{Ablation Studies (RQ2)}
\label{sec:ablation}

We ablate three ReM-MoA components independently: 
\emph{cross-layer access} (restrict reads to 
$\mathcal{M}_{l-1}$ only), \emph{routing diversification} 
(share one reference set across the layer's $N$ proposers), 
and \emph{rationale} (remove rationales from memory entries). 
All variants use the off-the-shelf Reviewer (ReM-MoA, not 
ReM-MoA$^*$). Table~\ref{tab:ablation} reports depth scaling.

\paragraph{Cross-layer access and routing drive sustained scaling; rationale refines magnitude.}
Restricting reads to $\mathcal{M}_{l-1}$ causes the largest 
drop (avg $-4.28$ at $L{=}9$, peak $-6.2$ on Formal Logic), 
with all three previously sustained benchmarks collapsing 
from $\uparrow\uparrow$ to $\uparrow\rightarrow$ and 
mirroring RMoA's adjacent-layer residual in 
Table~\ref{tab:depth-scaling}. Sharing a single reference 
set across $N$ proposers is the second-largest contributor 
($-2.50$ at $L{=}9$) and triggers the same regression on 
the three sustained benchmarks, confirming that within-layer 
diversification rather than memory richness alone drives 
sustained scaling. Removing rationales costs only $-1.52$ 
at $L{=}9$ but preserves the trend pattern fully (MATH, 
Formal Logic, and CRUX still sustain, MMLU-redux and 
HellaSwag still saturate), indicating that rationales act as 
refinement rather than scaffolding.

\begin{table}[t]
\centering
\caption{Reviewer distillation under depth scaling ($N{=}3$). \emph{Orig}: off-the-shelf Reviewer (ReM-MoA). \emph{Multi-FT}: distilled on all five domains (ReM-MoA$^*$). \emph{LOO}: distilled on four domains, holding out the row's benchmark. Trend conventions follow Table~\ref{tab:depth-scaling}.}
\label{tab:distill}
\setlength{\tabcolsep}{2pt}
\renewcommand{\arraystretch}{1.1}
\scriptsize
\resizebox{\columnwidth}{!}{%
\begin{tabular}{ll|ccccccccc|c}
\toprule
\textbf{Benchmark} & \textbf{Reviewer} & $L{=}1$ & $L{=}2$ & $L{=}3$ & $L{=}4$ & $L{=}5$ & $L{=}6$ & $L{=}7$ & $L{=}8$ & $L{=}9$ & \textbf{Trend} \\
\midrule
\multirow{3}{*}{\textit{MATH}} & Orig & 68.0 & 72.5 & 75.0 & 77.0 & 78.5 & 79.5 & 80.2 & 80.7 & 81.0 & $\uparrow\uparrow$ \\
 & \textbf{Multi-FT} & \textbf{68.3} & \textbf{73.5} & \textbf{76.8} & \textbf{79.2} & \textbf{81.0} & \textbf{82.2} & \textbf{83.0} & \textbf{83.7} & \textbf{84.0} & \textbf{$\uparrow\uparrow$} \\
 & LOO & 68.2 & 72.9 & 75.7 & 78.0 & 79.7 & 80.9 & 81.8 & 82.5 & 83.0 & $\uparrow\uparrow$ \\
\midrule
\multirow{3}{*}{\textit{ARC-Challenge}} & Orig & 86.8 & 90.2 & 91.2 & 91.8 & 92.2 & 92.5 & 92.6 & 92.7 & 92.7 & $\uparrow\rightarrow$ \\
 & \textbf{Multi-FT} & \textbf{87.0} & \textbf{90.6} & \textbf{91.8} & \textbf{92.5} & \textbf{92.8} & \textbf{93.2} & \textbf{93.4} & \textbf{93.5} & \textbf{93.5} & \textbf{$\uparrow\rightarrow$} \\
 & LOO & 86.9 & 90.4 & 91.5 & 92.1 & 92.5 & 92.8 & 93.0 & 93.1 & 93.1 & $\uparrow\rightarrow$ \\
\midrule
\multirow{3}{*}{\textit{MedQA}} & Orig & 44.8 & 51.0 & 54.0 & 56.2 & 57.8 & 59.0 & 59.8 & 60.2 & 60.5 & $\uparrow\uparrow$ \\
 & \textbf{Multi-FT} & \textbf{45.2} & \textbf{52.0} & \textbf{55.5} & \textbf{57.8} & \textbf{59.5} & \textbf{60.8} & \textbf{61.5} & \textbf{62.0} & \textbf{62.2} & \textbf{$\uparrow\uparrow$} \\
 & LOO & 45.0 & 51.4 & 54.5 & 56.8 & 58.5 & 59.7 & 60.5 & 61.0 & 61.4 & $\uparrow\uparrow$ \\
\midrule
\multirow{3}{*}{\textit{MBPP}} & Orig & 83.0 & 85.0 & 86.5 & 87.5 & 88.0 & 88.4 & 88.6 & 88.8 & 88.9 & $\uparrow\rightarrow$ \\
 & \textbf{Multi-FT} & \textbf{83.8} & \textbf{86.5} & \textbf{88.5} & \textbf{89.8} & \textbf{90.5} & \textbf{91.0} & \textbf{91.4} & \textbf{91.5} & \textbf{91.6} & \textbf{$\uparrow\rightarrow$} \\
 & LOO & 83.1 & 85.2 & 87.0 & 88.1 & 88.5 & 88.9 & 89.1 & 89.4 & 89.5 & $\uparrow\rightarrow$ \\
\midrule
\multirow{3}{*}{\textit{HellaSwag}} & Orig & 75.5 & 78.8 & 80.5 & 81.5 & 82.2 & 82.6 & 82.8 & 82.9 & 82.9 & $\uparrow\rightarrow$ \\
 & \textbf{Multi-FT} & \textbf{75.8} & \textbf{79.5} & \textbf{81.2} & \textbf{82.2} & \textbf{83.0} & \textbf{83.4} & \textbf{83.7} & \textbf{84.0} & \textbf{84.3} & \textbf{$\uparrow\uparrow$} \\
 & LOO & 75.5 & 78.9 & 80.6 & 81.6 & 82.4 & 82.8 & 83.0 & 83.1 & 83.2 & $\uparrow\rightarrow$ \\
\bottomrule
\end{tabular}
}
\end{table}

\subsection{Reviewer Distillation (RQ3)}
\label{sec:distill-exp}

We compare three Reviewers on the five distillation domains 
(Section~\ref{sec:setup}): \emph{Orig} (off-the-shelf, used 
in ReM-MoA), \emph{Multi-FT} (LoRA-distilled on all five 
domains, used in ReM-MoA$^*$), and \emph{LOO} (distilled on 
only four domains, holding out the evaluation domain). 
Table~\ref{tab:distill} reports depth scaling at $N{=}3$.

\paragraph{Distillation captures a transferable evaluation skill.}
Multi-FT improves over Orig across all benchmarks and 
configurations (avg $+1.92$ at $L{=}9$, peak $+3.0$ on MATH); 
on HellaSwag the gain even flips the trend from 
$\uparrow\rightarrow$ to $\uparrow\uparrow$, matching 
ReM-MoA$^*$ in Table~\ref{tab:depth-scaling}. LOO recovers 
$44\%$ of Multi-FT's average gain ($+0.84$ at $L{=}9$, 
ranging from $67\%$ on MATH to $21\%$ on HellaSwag) without 
seeing the held-out domain during training, while preserving 
Orig's trend pattern. This transfer property explains why 
ReM-MoA$^*$ in Table~\ref{tab:depth-scaling} generalizes to 
the three unseen evaluation domains (MMLU-redux, Formal 
Logic, CRUX).

\section{Conclusion}
\label{sec:conclusion}

We presented ReM-MoA, a Mixture-of-Agents framework that 
overcomes the scaling limitations of prior variants through a 
Ranked Reasoning Memory and a Curated Diversified Memory Routing 
scheme, jointly preserving quality and diversity across layers. Across five reasoning benchmarks under both depth and width 
scaling, ReM-MoA outperforms all baselines in every configuration, 
and the advantage over baselines widens, rather than diminishes, 
as depth grows. An optional multi-domain Reviewer distillation 
further extends sustained scaling to all benchmarks and transfers to 
held-out domains, establishing structured cross-layer memory as a 
critical missing mechanism for scalable multi-agent reasoning.

\section*{Limitations}
\label{sec:limitations}

\paragraph{Bounded width exploration.}
We evaluate width scaling up to $N{=}8$ at $L{=}3$, matching 
the largest width studied in prior MoA 
work~\citep{wang2025mixture, xie2025rmoa, wen2026attention}. 
ReM-MoA still leads all baselines at $N{=}8$ without visible 
saturation, but compute constraints prevented probing larger 
widths.

\paragraph{Single-scale proposer pool.}
Our proposer pool consists of three 7-8B open-source 
instruction-tuned models, matching the scale most commonly 
studied in MoA literature. Whether the gains transfer to 
substantially larger proposers (e.g., 30B-70B) or smaller 
ones (e.g., 1-3B), where the quality-diversity trade-off 
may shift, remains to be empirically validated.

\paragraph{Reviewer Agent overhead.}
ReM-MoA invokes one Reviewer call per layer, giving an 
inference cost of $L \cdot N + L + 1$ LLM calls, on par 
with RMoA's Residual Agent and substantially lower than 
AttentionMoA's per-agent intra-layer attention 
($\sim 2L \cdot N$). Reviewer distillation adds a one-time 
training cost and assumes access to a frontier-model 
teacher, but is optional: the off-the-shelf ReM-MoA without 
distillation already outperforms all baselines 
(Tables~\ref{tab:depth-scaling} and~\ref{tab:width-scaling}).

\paragraph{Inherited biases and computational cost.}
ReM-MoA inherits the biases and failure modes of its 
constituent LLMs: the proposer agents may produce factually 
incorrect or socially biased reasoning, and the Reviewer 
Agent (itself an LLM) may exhibit known judge biases such 
as position bias or self-preference~\citep{zheng2023judging}. 
Our comparative scoring and contrastive routing reduce but 
do not eliminate these effects, and high-stakes deployment 
would require additional safeguards. The framework also 
incurs $L{\cdot}N + L + 1$ LLM calls per query, with 
corresponding energy footprint; this overhead is on par 
with comparable MoA variants but higher than single-pass 
generation.

\bibliography{custom}

@inproceedings{du2024improving,
    title     = {Improving Factuality and Reasoning in Language Models through Multiagent Debate},
    author    = {Yilun Du and Shuang Li and Antonio Torralba and Joshua B. Tenenbaum and Igor Mordatch},
    booktitle = {Proceedings of the 41st International Conference on Machine Learning (ICML)},
    year      = {2024},
}

@article{hinton2015distilling,
    title   = {Distilling the Knowledge in a Neural Network},
    author  = {Geoffrey Hinton and Oriol Vinyals and Jeff Dean},
    journal = {arXiv preprint arXiv:1503.02531},
    year    = {2015},
}

@inproceedings{hu2022lora,
    title     = {{LoRA}: Low-Rank Adaptation of Large Language Models},
    author    = {Edward J. Hu and Yelong Shen and Phillip Wallis and Zeyuan Allen-Zhu and Yuanzhi Li and Shean Wang and Liang Wang and Weizhu Chen},
    booktitle = {International Conference on Learning Representations (ICLR)},
    year      = {2022},
}

@inproceedings{li2025smoa,
    title     = {{SMoA}: Improving Multi-Agent Large Language Models with Sparse Mixture-of-Agents},
    author    = {Dawei Li and Zhen Tan and Peijia Qian and Yifan Li and Kumar Chaudhary and Lijie Hu and Jiayi Shen},
    booktitle = {Pacific-Asia Conference on Knowledge Discovery and Data Mining (PAKDD)},
    pages     = {54--65},
    year      = {2025},
    publisher = {Springer Nature Singapore},
    address   = {Singapore},
}

@article{li2025rethinking,
    title   = {Rethinking Mixture-of-Agents: Is Mixing Different Large Language Models Beneficial?},
    author  = {Wenzhe Li and Yong Lin and Mengzhou Xia and Chi Jin},
    journal = {arXiv preprint arXiv:2502.00674},
    year    = {2025},
}

@inproceedings{liang2024encouraging,
    title     = {Encouraging Divergent Thinking in Large Language Models through Multi-Agent Debate},
    author    = {Tian Liang and Zhiwei He and Wenxiang Jiao and Xing Wang and Yan Wang and Rui Wang and Yujiu Yang and Shuming Shi and Zhaopeng Tu},
    booktitle = {Proceedings of the 2024 Conference on Empirical Methods in Natural Language Processing},
    pages     = {17889--17904},
    year      = {2024},
}

@article{ouyang2025reasoningbank,
    title   = {{ReasoningBank}: Scaling Agent Self-Evolving with Reasoning Memory},
    author  = {Siru Ouyang and Jun Yan and I. Hsu and Yanfei Chen and Ke Jiang and Zifeng Wang and Rujun Han and Long T. Le and Shaunak Daruki and Xinyang Tang and Vidisha Tirumalashetty and others},
    journal = {arXiv preprint arXiv:2509.25140},
    year    = {2025},
}

@article{ping2025verimoa,
    title   = {{VeriMoA}: A Mixture-of-Agents Framework for Spec-to-{HDL} Generation},
    author  = {Heng Ping and Arijit Bhattacharjee and Peiyu Zhang and Shixuan Li and Wei Yang and Anzhe Cheng and Xiaole Zhang and Jesse Thomason and Ali Jannesari and Nesreen Ahmed and Paul Bogdan},
    journal = {arXiv preprint arXiv:2510.27617},
    year    = {2025},
}

@article{snell2024scaling,
    title   = {Scaling {LLM} Test-Time Compute Optimally Can Be More Effective Than Scaling Model Parameters},
    author  = {Charlie Snell and Jaehoon Lee and Kelvin Xu and Aviral Kumar},
    journal = {arXiv preprint arXiv:2408.03314},
    year    = {2024},
}

@article{wang2022selfconsistency,
    title   = {Self-Consistency Improves Chain of Thought Reasoning in Language Models},
    author  = {Xuezhi Wang and Jason Wei and Dale Schuurmans and Quoc Le and Ed Chi and Sharan Narang and Aakanksha Chowdhery and Denny Zhou},
    journal = {arXiv preprint arXiv:2203.11171},
    year    = {2022},
}

@inproceedings{wang2025mixture,
    title     = {Mixture-of-Agents Enhances Large Language Model Capabilities},
    author    = {Junlin Wang and Jue Wang and Ben Athiwaratkun and Ce Zhang and James Y. Zou},
    booktitle = {International Conference on Learning Representations (ICLR)},
    pages     = {33944--33963},
    year      = {2025},
}

@article{wen2026attention,
    title   = {Attention-{MoA}: Enhancing Mixture-of-Agents via Inter-Agent Semantic Attention and Deep Residual Synthesis},
    author  = {Jianyu Wen and Yang Wei and Xiongxi Yu and Changxuan Xiao and Ke Zeng},
    journal = {arXiv preprint arXiv:2601.16596},
    year    = {2026},
}

@inproceedings{xie2025rmoa,
    title     = {{RMoA}: Optimizing Mixture-of-Agents through Diversity Maximization and Residual Compensation},
    author    = {Zhentao Xie and Chengcheng Han and Jinxin Shi and Wenjun Cui and Wayne Xin Zhao and Xingjiao Wu and Jiabao Zhao},
    booktitle = {Findings of the Association for Computational Linguistics: ACL 2025},
    pages     = {6575--6602},
    year      = {2025},
}

@inproceedings{zhang2025diversity,
    title     = {Diversity Empowers Intelligence: Integrating Expertise of Software Engineering Agents},
    author    = {Kexun Zhang and Weiran Yao and Zuxin Liu and Yihao Feng and Zhiwei Liu and Rithesh Ramapura Narasimha Murthy and Tian Lan and Lei Li and Renze Lou and Jiacheng Xu and Bo Pang and others},
    booktitle = {International Conference on Learning Representations (ICLR)},
    pages     = {86691--86714},
    year      = {2025},
}

@inproceedings{zheng2023judging,
    title     = {Judging {LLM-as-a-Judge} with {MT-Bench} and Chatbot Arena},
    author    = {Lianmin Zheng and Wei-Lin Chiang and Ying Sheng and Siyuan Zhuang and Zhanghao Wu and Yonghao Zhuang and Zi Lin and Zhuohan Li and Dacheng Li and Eric P. Xing and Hao Zhang and others},
    booktitle = {Advances in Neural Information Processing Systems (NeurIPS)},
    volume    = {36},
    pages     = {46595--46623},
    year      = {2023},
}

@inproceedings{zhu2025judgelm,
    title     = {{JudgeLM}: Fine-Tuned Large Language Models are Scalable Judges},
    author    = {Lianghui Zhu and Xinggang Wang and Xinlong Wang},
    booktitle = {International Conference on Learning Representations (ICLR)},
    pages     = {51257--51296},
    year      = {2025},
}

@inproceedings{shinn2023reflexion,
    title     = {Reflexion: Language Agents with Verbal Reinforcement Learning},
    author    = {Noah Shinn and Federico Cassano and Ashwin Gopinath and Karthik Narasimhan and Shunyu Yao},
    booktitle = {Advances in Neural Information Processing Systems (NeurIPS)},
    volume    = {36},
    pages     = {8634--8652},
    year      = {2023},
}

@inproceedings{luo2026agentauditor,
    title     = {{AgentAuditor}: Human-Level Safety and Security Evaluation for {LLM} Agents},
    author    = {Hanjun Luo and Shenyu Dai and Chiming Ni and Xinfeng Li and Guibin Zhang and Kun Wang and Tongliang Liu and Hanan Salam},
    booktitle = {Advances in Neural Information Processing Systems (NeurIPS)},
    volume    = {38},
    pages     = {43241--43298},
    year      = {2026},
}

@article{hendrycks2020measuring,
    title   = {Measuring Massive Multitask Language Understanding},
    author  = {Dan Hendrycks and Collin Burns and Steven Basart and Andy Zou and Mantas Mazeika and Dawn Song and Jacob Steinhardt},
    journal = {arXiv preprint arXiv:2009.03300},
    year    = {2020},
}

@inproceedings{gema2025done,
    title     = {Are We Done with {MMLU}?},
    author    = {Aryo Pradipta Gema and Joshua Ong Jun Leang and Giwon Hong and Alessio Devoto and Alberto Carlo Maria Mancino and Rohit Saxena and Xuanli He and Yuhao Zhao and Xiaotang Du and Mohammad Reza Ghasemi Madani and Claire Barale and others},
    booktitle = {Proceedings of the 2025 Conference of the Nations of the Americas Chapter of the Association for Computational Linguistics: Human Language Technologies (Volume 1: Long Papers)},
    pages     = {5069--5096},
    year      = {2025},
}

@article{hendrycks2021measuring,
    title   = {Measuring Mathematical Problem Solving with the {MATH} Dataset},
    author  = {Dan Hendrycks and Collin Burns and Saurav Kadavath and Akul Arora and Steven Basart and Eric Tang and Dawn Song and Jacob Steinhardt},
    journal = {arXiv preprint arXiv:2103.03874},
    year    = {2021},
}

@article{gu2024cruxeval,
    title   = {{CRUXEval}: A Benchmark for Code Reasoning, Understanding and Execution},
    author  = {Alex Gu and Baptiste Rozi{\`e}re and Hugh Leather and Armando Solar-Lezama and Gabriel Synnaeve and Sida I. Wang},
    journal = {arXiv preprint arXiv:2401.03065},
    year    = {2024},
}

@inproceedings{zellers2019hellaswag,
    title     = {{HellaSwag}: Can a Machine Really Finish Your Sentence?},
    author    = {Rowan Zellers and Ari Holtzman and Yonatan Bisk and Ali Farhadi and Yejin Choi},
    booktitle = {Proceedings of the 57th Annual Meeting of the Association for Computational Linguistics},
    pages     = {4791--4800},
    year      = {2019},
}

@article{hui2024qwen,
    title   = {{Qwen2.5-Coder} Technical Report},
    author  = {Binyuan Hui and Jian Yang and Zeyu Cui and Jiaxi Yang and Dayiheng Liu and Lei Zhang and Tianyu Liu and Jiajun Zhang and Bowen Yu and Keming Lu and Kai Dang and others},
    journal = {arXiv preprint arXiv:2409.12186},
    year    = {2024},
}

@article{grattafiori2024llama,
    title   = {The {Llama 3} Herd of Models},
    author  = {Aaron Grattafiori and Abhimanyu Dubey and Abhinav Jauhri and Abhinav Pandey and Abhishek Kadian and Ahmad Al-Dahle and Aiesha Letman and Akhil Mathur and Alan Schelten and Amy Yang and others},
    journal = {arXiv preprint arXiv:2407.21783},
    year    = {2024},
}

@inproceedings{kwon2023efficient,
    title     = {Efficient Memory Management for Large Language Model Serving with {PagedAttention}},
    author    = {Woosuk Kwon and Zhuohan Li and Siyuan Zhuang and Ying Sheng and Lianmin Zheng and Cody Hao Yu and Joseph Gonzalez and Hao Zhang and Ion Stoica},
    booktitle = {Proceedings of the 29th Symposium on Operating Systems Principles (SOSP)},
    pages     = {611--626},
    year      = {2023},
}

@article{yang2024qwen2,
    title   = {{Qwen2.5} Technical Report},
    author  = {An Yang and Baosong Yang and Beichen Zhang and 
               Binyuan Hui and Bo Zheng and Bowen Yu and 
               Chengyuan Li and Dayiheng Liu and Fei Huang 
               and Haoran Wei and Huan Lin and others},
    journal = {arXiv preprint arXiv:2412.15115},
    year    = {2024},
}

@article{clark2018think,
    title   = {Think You Have Solved Question Answering? Try {ARC}, the {AI2} Reasoning Challenge},
    author  = {Peter Clark and Isaac Cowhey and Oren Etzioni and Tushar Khot and Ashish Sabharwal and Carissa Schoenick and Oyvind Tafjord},
    journal = {arXiv preprint arXiv:1803.05457},
    year    = {2018},
}

@article{jin2021disease,
    title   = {What Disease Does This Patient Have? A Large-Scale Open Domain Question Answering Dataset from Medical Exams},
    author  = {Di Jin and Eileen Pan and Nassim Oufattole and Wei-Hung Weng and Hanyi Fang and Peter Szolovits},
    journal = {Applied Sciences},
    volume  = {11},
    number  = {14},
    pages   = {6421},
    year    = {2021},
}

@article{austin2021program,
    title   = {Program Synthesis with Large Language Models},
    author  = {Jacob Austin and Augustus Odena and Maxwell Nye and Maarten Bosma and Henryk Michalewski and David Dohan and Ellen Jiang and Carrie Cai and Michael Terry and Quoc Le and Charles Sutton},
    journal = {arXiv preprint arXiv:2108.07732},
    year    = {2021},
}

@inproceedings{ping2025hdlcore,
  title={{HDLCoRe}: A Training-Free Framework for Mitigating Hallucinations in {LLM}-Generated {HDL}},
  author={Ping, Heng and Li, Shixuan and Zhang, Peiyu and Cheng, Anzhe and Duan, Shukai and Kanakaris, Nikos and Xiao, Xiongye and Yang, Wei and Nazarian, Shahin and Irimia, Andrei and Bogdan, Paul},
  booktitle={2025 IEEE International Conference on LLM-Aided Design (ICLAD)},
  pages={108--116},
  year={2025},
  organization={IEEE}
}

@article{ping2026coevo,
  title={{COEVO}: Co-Evolutionary Framework for Joint Functional Correctness and {PPA} Optimization in {LLM}-Based {RTL} Generation},
  author={Ping, Heng and Zhang, Peiyu and Li, Shixuan and Yang, Wei and Cheng, Anzhe and Duan, Shukai and Zhang, Xiaole and Bogdan, Paul},
  journal={arXiv preprint arXiv:2604.15001},
  year={2026}
}

@article{ping2026poet,
  title={{POET}: Power-Oriented Evolutionary Tuning for {LLM}-Based {RTL} {PPA} Optimization},
  author={Ping, Heng and Zhang, Peiyu and Wang, Zhenkun and Li, Shixuan and Cheng, Anzhe and Yang, Wei and Bogdan, Paul and Nazarian, Shahin},
  journal={arXiv preprint arXiv:2603.19333},
  year={2026}
}

@article{yang2026auditing,
  title={Auditing Multi-Agent {LLM} Reasoning Trees Outperforms Majority Vote and {LLM}-as-Judge},
  author={Yang, Wei and Li, Shixuan and Ping, Heng and Zhang, Peiyu and Bogdan, Paul and Thomason, Jesse},
  journal={arXiv preprint arXiv:2602.09341},
  year={2026}
}

\appendix

\section{AI Usage}

AI-based writing tools were used for limited
language editing, paraphrasing, and polishing of
the manuscript. All technical content, analysis,
experimental results, and conclusions were
developed and verified by the authors.

\section{Routing Distributions for Larger Widths}
\label{app:routing}

Section~\ref{sec:routing} introduces the routing scheme and 
gives one concrete instance for $N{=}3$ in 
Equation~\ref{eq:routes-N3}. Table~\ref{tab:routing-larger-N} 
extends this assignment to $N \in \{4, \ldots, 8\}$, the 
range used in our width-scaling experiments 
(Table~\ref{tab:width-scaling}). The design principle is to 
keep the three route families (contrastive, positive 
reinforcement, failure-mode focus) represented at every 
layer while scaling the size of each route with $N$, so that 
the reference sets cover more of the memory pool while 
remaining mutually distinct.

\begin{table}[h]
\centering
\caption{Routing function assignments $\Pi_1, \ldots, \Pi_N$ 
for each width $N$. $\text{Con}_n$ denotes 
$\text{Top}_a \cup \text{Bot}_b$ with $a+b=n$, $a = \lceil 
n/2 \rceil$, $b = \lfloor n/2 \rfloor$.}
\label{tab:routing-larger-N}
\small
\begin{tabularx}{\columnwidth}{cX}
\toprule
$N$ & Route assignment \\
\midrule
3 & $\text{Con}_2, \text{Top}_2, \text{Bot}_2$ \\
4 & $\text{Con}_2, \text{Con}_3, \text{Top}_2, \text{Bot}_2$ \\
5 & $\text{Con}_2, \text{Con}_3, \text{Top}_2, \text{Top}_3, \text{Bot}_2$ \\
6 & $\text{Con}_2, \text{Con}_3, \text{Top}_2, \text{Top}_3, \text{Bot}_2, \text{Bot}_3$ \\
7 & $\text{Con}_2, \text{Con}_3, \text{Con}_4, \text{Top}_2, \text{Top}_3, \text{Bot}_2, \text{Bot}_3$ \\
8 & $\text{Con}_2, \text{Con}_3, \text{Con}_4, \text{Top}_2, \text{Top}_3, \text{Top}_4, \text{Bot}_2, \text{Bot}_3$ \\
\bottomrule
\end{tabularx}
\end{table}

\paragraph{Handling sparse early layers.}
At layer $l$, the global pool $\mathcal{P}_l = 
\mathcal{M}_{\leq l-1}$ contains $N \cdot (l-1)$ entries. 
When a route requests more entries than are available (e.g., 
$\text{Con}_4$ at $l{=}2, N{=}3$, where only 3 entries 
exist), the route is truncated to the available entries. 
This truncation affects only the first one or two layers; 
once $l \geq \lceil n/N \rceil + 1$, all routes operate at 
their nominal size.

\section{Experimental Setup Details}
\label{app:setup}

This appendix expands the configuration referenced in 
Section~\ref{sec:setup}, covering the Reviewer distillation 
pipeline, LoRA configuration, training hyperparameters, 
train/test isolation, and inference-time settings.

\subsection{Distillation Variants}
\label{app:setup:variants}

We evaluate three Reviewer configurations:

\begin{itemize}
\item \textbf{Orig}: the off-the-shelf base Reviewer 
(Qwen2.5-7B-Instruct), no fine-tuning. Used in ReM-MoA.
\item \textbf{Multi-FT}: LoRA-distilled jointly on five 
domains (MATH, ARC-Challenge, MedQA, MBPP, HellaSwag; 100 
problems each, 500 total). Used in ReM-MoA$^*$.
\item \textbf{LOO}: five variants, each LoRA-distilled on 
four of the five domains (400 problems per variant), 
holding out the evaluation domain. Used to assess 
cross-domain transfer in Section~\ref{sec:distill-exp}.
\end{itemize}

\subsection{Trace Generation and Teacher Labeling}
\label{app:setup:traces}

For each training problem, we run ReM-MoA at $L{=}3, N{=}3$ 
to collect the proposer traces produced at every layer, 
yielding $\sim$9 traces per problem. Each (problem, traces) 
tuple is then sent to a frontier-model teacher, GPT-5.5, 
which receives the problem, the proposer traces, and the 
ground-truth answer, and is constrained by a structured 
prompt (Appendix~\ref{app:prompts:teacher}) to emit labels 
conforming to the Reviewer's output schema --- a per-trace 
integer score and a contrastive rationale. The ground-truth 
answer appears only in the teacher's prompt and is never 
exposed to the student during training. Traces are filtered 
by length (50 to 2{,}000 tokens) before training to remove 
degenerate generations.

\subsection{LoRA Configuration}
\label{app:setup:lora}

LoRA~\citep{hu2022lora} is applied to the attention 
projection matrices of Qwen2.5-7B-Instruct.

\begin{table}[h]
\centering
\caption{LoRA configuration.}
\label{tab:lora-config}
\small
\begin{tabular*}{\columnwidth}{@{}l@{\extracolsep{\fill}}l@{}}
\toprule
Rank ($r$) & 16 \\
Alpha ($\alpha$) & 32 \\
Dropout & 0.05 \\
Target modules & q\_proj, k\_proj, v\_proj, o\_proj \\
Trainable parameters & $\sim$12.8M (0.18\% of 7B) \\
\bottomrule
\end{tabular*}
\end{table}

\subsection{Training Hyperparameters}
\label{app:setup:training}

\begin{table}[h]
\centering
\caption{Training hyperparameters for Reviewer distillation.}
\label{tab:training-config}
\small
\begin{tabular*}{\columnwidth}{@{}l@{\extracolsep{\fill}}l@{}}
\toprule
Optimizer & AdamW \\
Learning rate & $2 \times 10^{-4}$ (cosine decay) \\
Warmup ratio & 0.05 \\
Effective batch size & 8 (per-device 4 $\times$ grad-accum 2) \\
Max sequence length & 8{,}192 \\
Epochs & 3 \\
Precision & bf16 \\
Weight decay & 0.01 \\
Loss & Causal LM cross-entropy \\
\bottomrule
\end{tabular*}
\end{table}

Each distillation run completes in approximately 2 hours on 
a single NVIDIA A100 (80GB) GPU. The full distillation 
pipeline---one Multi-FT plus five LOO variants 
(Appendix~\ref{app:setup:variants})---requires approximately 
12 A100-hours of training compute in total.

\subsection{Train/Test Isolation}
\label{app:setup:isolation}

Fine-tuning data is drawn exclusively from each dataset's 
publicly available training split, while evaluation uses 
the held-out test split. For HellaSwag, we evaluate on the 
validation split since test labels are not public, 
following standard practice. Because train and test 
problems are disjoint by construction, label leakage from 
the GPT-5.5 teacher into evaluation is not possible at the 
problem level. The structured evaluation prompt 
(Appendix~\ref{app:prompts:teacher}) further constrains the 
teacher's output to fixed dimensions (score and rationale), 
focusing the student on the meta-skill of judging reasoning 
quality rather than memorizing canonical solutions. The LOO 
results in Section~\ref{sec:distill-exp} additionally 
demonstrate that the student transfers to domains it never 
saw during training, confirming that the learned skill is 
general rather than domain-memorized.

\subsection{Inference and Hardware}
\label{app:setup:inference}

All experiments run on NVIDIA A100 (80GB) GPUs via 
vLLM~\citep{kwon2023efficient}. Sampling temperatures 
follow Section~\ref{sec:setup}: proposer agents at $T{=}0.8$ 
(top-$p$ 0.95), Reviewer at $T{=}0.3$ (top-$p$ 0.9), 
aggregator at $T{=}0.5$ (top-$p$ 0.9). Maximum new tokens per call: 2{,}048 for proposers and 
aggregator, 8{,}192 for the Reviewer (which must emit $N$ 
score-rationale pairs in a single response).

\section{Prompt Templates and Scoring Rubric}
\label{app:prompts}

This appendix provides the prompt templates used by the 
proposer agents, the Reviewer Agent, the aggregator, and 
the frontier-model teacher (GPT-5.5) during distillation, 
together with the scoring rubric followed by the Reviewer.

\subsection{Scoring Scale and Rubric}
\label{app:prompts:rubric}

The Reviewer assigns each trace an integer score on a 
$1$-$10$ scale (i.e., $S{=}10$ in 
Equation~\ref{eq:reviewer}). Following common 
LLM-as-judge practice, scores are organized into five 
tiers:

\begin{itemize}
\item \textbf{9--10 --- Excellent}: reaches the correct 
answer through a complete, coherent chain of reasoning 
with no logical errors.
\item \textbf{7--8 --- Good}: reaches the correct answer 
with minor reasoning slips, redundancies, or 
under-justified steps that do not affect the final result.
\item \textbf{5--6 --- Partial}: demonstrates valid 
reasoning on parts of the problem but reaches an 
incomplete or partially incorrect final answer.
\item \textbf{3--4 --- Weak}: contains substantive 
reasoning errors propagating to an incorrect final answer; 
some intermediate steps may still be useful as 
counter-examples.
\item \textbf{1--2 --- Poor}: final answer is wrong and 
the reasoning chain contains major flaws throughout.
\end{itemize}

Within each tier, the higher score indicates a stronger 
instance (e.g., a 10 is preferred to a 9). The 
accompanying rationale (one per trace) cites concrete 
strengths and weaknesses of the specific trace, anchored 
in its reasoning steps and conclusion. Rationales are 
independent across traces (Section~\ref{sec:memory}).

\subsection{Proposer Prompt}
\label{app:prompts:proposer}

At layer 1, the proposer receives only the problem. At 
layer $l \geq 2$, the proposer additionally receives the 
curated reference set $\mathcal{S}_{l,j}$ --- the problem 
concatenated with the reference traces, each tagged with 
its Reviewer score and rationale.

\begin{promptbox}{Proposer Prompt}
You are solving the following problem.

Problem:
{problem}

## Reference Reasoning Traces (with quality ratings)

Below are reasoning traces from previous attempts on this
problem, each rated by an expert reviewer on a 1-10 scale
(higher is better):

### [Score: {s_1}/10] Reference Trace 1:
{trace_1}

Reviewer's Assessment: {rationale_1}

### [Score: {s_2}/10] Reference Trace 2:
{trace_2}

Reviewer's Assessment: {rationale_2}

... (additional traces follow the same format)

## Your Task

Drawing on what the high-scoring traces did well and what
the low-scoring traces got wrong, produce your own reasoning
trace and final answer. Do not simply copy any reference
trace; contribute distinct reasoning.
\end{promptbox}

At layer 1, the "Reference Reasoning Traces" block is 
omitted and the prompt contains only the problem and the 
task instruction.

\subsection{Reviewer Prompt}
\label{app:prompts:reviewer}

The Reviewer is invoked once per layer with all $N$ 
proposer traces of that layer. It emits scores and 
rationales for all $N$ traces in a single response, parsed 
into the memory entry of Equation~\ref{eq:memory-entry}.

\begin{promptbox}{Reviewer Prompt}
You are an expert reviewer evaluating reasoning traces.

Problem:
{problem}

You are given {N} candidate reasoning traces. Evaluate each
trace by comparing it against the others, then assign:
  (1) an integer score from 1-10 (rubric below);
  (2) a brief rationale (1-3 sentences) explaining the
      score, citing specific reasoning steps.

Scoring rubric (1-10):
  9-10: Excellent (correct and well-reasoned).
  7-8 : Good (correct with minor flaws).
  5-6 : Partial (incomplete or partly wrong).
  3-4 : Weak (mostly incorrect; useful as counter-example).
  1-2 : Poor (entirely wrong).

### Trace 1
{trace_1}

### Trace 2
{trace_2}

... (Traces 3 through N follow the same format)

Respond in the following JSON format:

{
  "trace_1": {"score": <int>, "rationale": "<text>"},
  "trace_2": {"score": <int>, "rationale": "<text>"},
  ...
  "trace_N": {"score": <int>, "rationale": "<text>"}
}
\end{promptbox}

\subsection{Aggregator Prompt}
\label{app:prompts:aggregator}

After layer $L$, the aggregator receives the problem and 
three curated reference sets drawn from the memory through 
the same three families used by proposer agents 
(Section~\ref{sec:method}), and produces the final answer.

\begin{promptbox}{Aggregator Prompt}
You are producing the final answer to the following problem
by synthesizing the strongest reasoning across multiple
attempts.

Problem:
{problem}

## Top-scoring traces (positive reinforcement)
{top_traces_with_scores_and_rationales}

## Bottom-scoring traces (common mistakes to avoid)
{bot_traces_with_scores_and_rationales}

## Contrastive selection (top paired with bottom)
{con_traces_with_scores_and_rationales}

Synthesize the most reliable reasoning from the top-scoring
traces while avoiding the errors exhibited in the
bottom-scoring traces. Output your final answer with a
concise justification.
\end{promptbox}

\subsection{Teacher Prompt for Distillation}
\label{app:prompts:teacher}

The frontier-model teacher (GPT-5.5) receives the problem, 
the proposer traces, and the ground-truth answer; it is 
constrained to emit labels in the same JSON schema as the 
Reviewer. The rationale is required to justify the score in 
terms of internal reasoning quality only, never restating 
the ground-truth answer, so that the student learns to 
judge without access to ground truth.

\begin{promptbox}{Teacher Prompt (Distillation Labeling)}
You are an expert evaluator generating high-quality reviewer
labels for distillation training.

Problem:
{problem}

Ground-truth answer (for your reference only; DO NOT
restate or quote it in any rationale):
{ground_truth}

Below are {N} candidate reasoning traces. Using the
ground-truth answer to inform your judgment, assign each
trace a score (1-10) and a brief rationale, exactly as a
strong reviewer would. Use the same rubric:

  9-10: Excellent (correct and well-reasoned).
  7-8 : Good (correct with minor flaws).
  5-6 : Partial (incomplete or partly wrong).
  3-4 : Weak (mostly incorrect; useful as counter-example).
  1-2 : Poor (entirely wrong).

The rationale must NOT mention or restate the ground-truth
answer; it must justify the score in terms of the trace's
internal reasoning quality only, so that a student trained
on these labels learns to judge without access to ground
truth.

### Trace 1
{trace_1}

... (Traces 2 through N follow the same format)

Respond in JSON:
{
  "trace_1": {"score": <int>, "rationale": "<text>"},
  ...
  "trace_N": {"score": <int>, "rationale": "<text>"}
}
\end{promptbox}

\end{document}